\documentclass{bmvc2k}

\usepackage{makecell}
\usepackage{xcolor}

\title{Differentiable SLAM Helps Deep Learning-based LiDAR Perception Tasks
}

\addauthor{Prashant Kumar}{prahantk.nan@gmail.com}{1}
\addauthor{Dheeraj Vattikonda}{dheeraj.46329@gmail.com}{2}
\addauthor{Vedang Bhupesh Shenvi Nadkarni}{ vedang.nadkarni@gmail.com}{3}
\addauthor{Erqun Dong}{ doneq13@gmail.com}{2}
\addauthor{Sabyasachi Sahoo}{sabyasachi.sahoo.1@ulaval.ca}{4}
\addinstitution{
 IIT Delhi, India
}
\addinstitution{
 McGill University, MILA
}
\addinstitution{
 BITS Pilani, India
}
\addinstitution{
 Université Laval, MILA
}

\runninghead{Kumar et. al.}{Diff-SLAM Helps DL-based LiDAR Perception Tasks}

\begin{document}

\maketitle

\begin{abstract}
We investigate a new paradigm  that uses differentiable SLAM architectures in a self-supervised manner to train end-to-end deep learning models in various LiDAR based applications. To the best of our knowledge there does not exist any work that leverages  SLAM as a training signal for deep learning based models. We explore new ways to improve the efficiency, robustness, and adaptability of LiDAR systems with deep learning techniques. We focus on the potential benefits of differentiable SLAM architectures for improving performance of deep learning tasks such as classification, regression as well as SLAM. Our experimental results demonstrate a non-trivial increase in the performance of two deep learning applications - Ground Level Estimation and Dynamic to Static LiDAR Translation, when used with differentiable SLAM architectures. Overall, our findings provide important insights that  enhance the performance of LiDAR based navigation systems. We demonstrate that this new paradigm of using SLAM Loss signal while training LiDAR based models can be easily adopted by the community.
\end{abstract} 
\section{Introduction}
\label{sec:intro}
We investigate the impact of differentiable SLAM on training better deep learning-based machine perception (DLMP) models through the methodology of fully differentiable backpropagation.
SLAM is a foundational component in mobile robotics. Robots build a map of their environment while simultaneously determining their locations within the map.
The potential of using SLAM to help DLMP has been used on tasks such as learning observation models for new modalities~\cite{sodhi2022leo}, object localization and tracking~\cite{merrill2022symmetry, lu2022slam}, etc. 
State-of-the-art SLAM systems are often not differentiable, presenting a challenge in integrating them with deep learning approaches.
While recent works such as GradSLAM~\cite{gradslam} and GradLidarSLAM~\cite{fnu2022gradlidarslam} have addressed this issue by proposing differentiable SLAM architectures, there has been little investigation into how these architectures impact the performance of deep learning models. 
It is an open question whether differentiable SLAM architectures can be effectively utilized to enhance the performance of deep learning models in various LiDAR based applications.
We propose a self-supervised framework that leverages a differentiable SLAM architecture. It enable fully differentiable training of deep learning models with SLAM error for various LiDAR applications. Our method is based on the principle of minimizing the discrepancy between the output of the deep learning model and the ground truth, including the trajectory error obtained from ground truth and predicted LiDAR scans. Through extensive experimentation, we demonstrate that our approach outperforms existing methods and achieves improvements in deep learning tasks. Our results highlight the potential of utilizing differentiable SLAM architectures to enhance the performance of deep learning models. Our main contributions are: 

\begin{itemize}
    \item We propose a new framework to train differentiable LiDAR-based SLAM using deep learning-based machine perception (DLMP) tasks in a self-supervised manner.
    \item We demonstrate its effectiveness by applying it on two tasks - (1) Ground Plane Estimation and Ground Point Segmentation (2) Dynamic to static LiDAR translation for improved SLAM.
    \item Our experiments show that our proposed framework significantly improves the performance of deep learning models on DLMP tasks.
\end{itemize}
\subsection{Related Work}
\subsubsection{Differentiable SLAM}
The idea of making SLAM differentiable has been investigated in some previous works~\cite{gradslam,fnu2022gradlidarslam,yi2021differentiable}.
Incorporating differentiable SLAM modules to help deep learning training has huge potential. However, to the best of our knowledge it has not been implemented and well studied.
Several works on SLAM that integrate deep learning-based techniques have been introduced in recent years - learning observation model for new modalities~\cite{sodhi2022leo}, learning object pose tracking~\cite{lu2022slam,merrill2022symmetry}, learning a compact scene representation~\cite{bloesch2018codeslam,Zhi2019SceneCodeMD}, learning a CNN-based depth predictor as the front-end of a monocular SLAM system~\cite{Tateno2017CNNSLAMRD}, etc.
While these works leverage learning techniques, they often focus only on specific modules within the SLAM system. Furthermore, these methods are typically limited to visual odometry.
Sodhi et.a l. \cite{sodhi2022leo} attempt to optimize end-to-end tracking performance by learning observation models using energy-based methods for SLAM on novel modalities like tactile sensors. They do not use trajectory error directly to optimize the observation model. They do not conduct perception tasks explicitly (i.e. no perception based results are available).
Different from existing literature, we investigate the use of SLAM trajectory error in a fully differentiable fashion to help LiDAR based deep learning tasks.
\subsubsection{LiDAR based Deep Learning}
\label{related-2}
Several works \cite{caccia2018deep, nakashima2021learning,kim2020color,zyrianov2022learning,triess2022realism,nakashima2023generative,eskandar2022glpu,guillard2022learning} have explored generative modelling for LiDAR. 
LiDAR based generative modelling was first introduced by Caccia et. al. \cite{caccia2018deep}. They use deep generative models - VAE, GANs to reconstruct as well as generate high quality LiDAR samples. Another work, DSLR~\cite{kumar2021dynamic}, extended this idea to generate static structures occluded by dynamic objects for 3D LiDAR scene reconstruction in an adversarial setting. The work also aims to improve SLAM performance with these static reconstructions. We therefore consider DSLR as a suitable test bed for our work. Another work \cite{nakashima2021learning} focuses on alleviating the problem of dropped points on the LiDAR depth map by introducing measurement uncertainty in the generative models. 
LiDAR data is a rich source of information for the 3D world of vital use for autonomous navigation systems. There exists a good body of work on LiDAR based segmentation \cite{milioto2019rangenet++,zhang2020polarnet,qi2018frustum,landrieu2018large,chen2021moving,landrieu2018large,hu2020randla,bloembergen2021automatic}, object detection \cite{lang2019pointpillars,zhu2020ssn,he2020structure,yan2018second,yang20203dssd,shi2019pointrcnn,shi2020points,yin2021center}, ground elevation estimation \cite{paigwar2020gndnet,chen2014automatic,lim2021patchwork,lee2022patchwork++}. 
For tasks other than generative modelling, one of the requirements for utilizing differentiable SLAM based error is that the output must be in the form of a per point prediction/regression so that it can be mapped to the original LiDAR points. A subset of points is then selected for SLAM. SLAM can then be performed between the mapped predicted LiDAR and the input LiDAR for SLAM Loss back-propagation. However, multi-class (except binary) per-point classification based tasks(e.g. segmentation) require certain non differentiable operations (e.g. \verb|torch.isin(), torch.argmax()|, etc) to map the predictions to the original LiDAR based on a given criteria (e.g. only static object classes) and cannot be integrated with differentiable SLAM. This is a limitation of the  differentiable SLAM module.

We choose a task that unifies binary classification as well as regression to show the effect of  differentiable SLAM on the selected task. Ground plane estimation and ground point segmentation unifies both these modalities. We use this task to show the benefit of differentiable SLAM. We use a well-known baseline GndNet \cite{paigwar2020gndnet} that has shown impressive performance on the above mentioned task.

Our primary focus is to propose a highly accurate SLAM solution that can provide more effective supervisory signals. Current DL-based supervised pose estimation  methods may introduce errors into deep learning perception models, while differentiable SLAM has shown promising results and offers better accuracy for our task. Therefore we do not currently use DL-based pose estimation methods for our work.
\begin{figure*}[]
\begin{subfigure}{\textwidth 0pt}
  \centering
  \includegraphics[width=1\linewidth, height=0.4\linewidth]{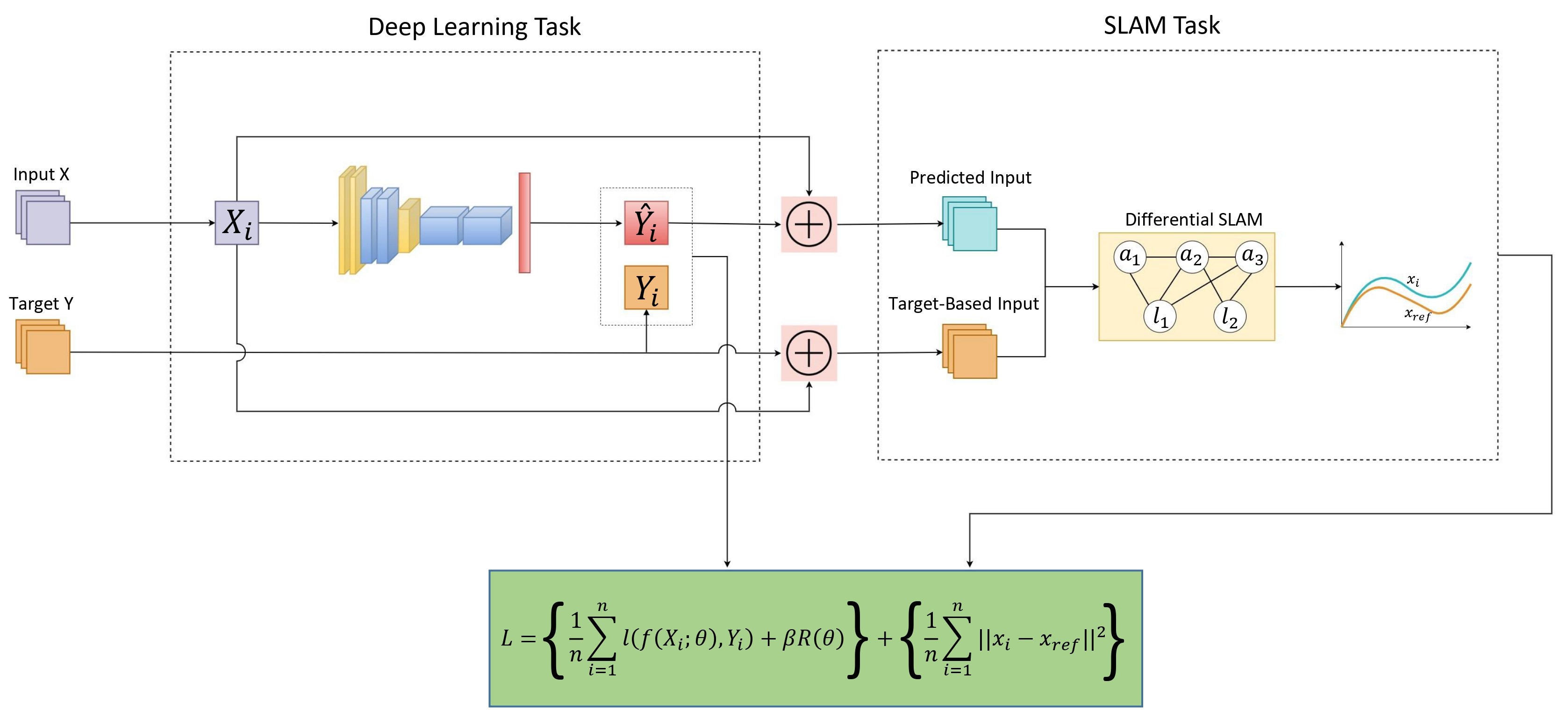} \caption{Our proposed framework for integrating differentiable SLAM into deep learning task training. The framework includes a Deep Learning Task which takes the input scans ($X_i$) and outputs the predicted scans ($\hat{Y}_i$). In the SLAM task, the trajectory error is calculated between trajectories $x_i$(calculated from predicted scans $\hat{Y}_i$) and $x_{ref}$(calculated from target scans $Y_i$). The framework aims to optimize the combined loss of both Deep Learning loss and SLAM (Absolute Trajectory Loss between trajectories) Loss. }
  \label{fig:sub-first}
\end{subfigure}
\end{figure*}
\section{Problem Setup}
\subsection{Model}
Our framework is composed of two primary components: a generic deep learning module and a differentiable SLAM module. %
These are coupled together to allow training of the entire architecture in an end-to-end fashion. We focus on optimizing the overall loss function - the sum of the loss for the deep learning model and the SLAM module, denoted as
\begin{equation}
Loss = L_{model} + \gamma L_{slam},
\end{equation}
where $\gamma$ is a coefficient balancing the impact of the SLAM on deep learning.
The loss function for the deep learning model $L_{model}$ is defined as
\begin{equation}
L_{model} = \frac{1}{n} \sum_{i=1}^{n} l(f(X_i; \theta), Y_i) + \beta R(\theta),
\end{equation}
where $n$ is the number of training samples, $f$ is the deep learning model with parameters $\theta$, $l$ is a loss function, $X_i$ is the input to the model, $Y_i$ is the ground truth output, $\beta$ is a regularization parameter, and $R(\theta)$ is the regularization term.

The SLAM loss $L_{slam}$ includes both translational and orientational errors. These are used to calculate the SLAM loss between the trajectories, which is defined as
\begin{equation}
L_{slam} = \frac{1}{n} \sum_{i=1}^{n} ||x_i - x_{ref}||^2,
\end{equation}
 $x_i$ is the estimated way-point of the trajectory generated by the SLAM algorithm using the predicted output of the deep learning model. $x_{ref}$ is the estimated ground truth way-point, generated by estimating the trajectory from the input LiDAR sequence instead of using actual ground-truth pose estimates that are available in the dataset. This enables the differential SLAM framework to work in a self-supervised fashion. It makes our method practically viable and efficient.
\subsection{Learning with SLAM}
To enhance the performance of deep learning by optimizing SLAM errors, it is necessary for the SLAM to be fully differentiable.
Our differentiable SLAM module takes two branches of input and predicts one trajectory for each of them.
First - the outputs of the deep learning model (e.g. generated LiDAR scan, predicted segmentation mask) are input to the differentiable SLAM as input and the differentiable SLAM predicts a trajectory based on deep learning's output.
Second - ground truth LiDAR information (e.g. LiDAR scan with only static points annotated, ground truth segmentation masks) are input to the differentiable SLAM, and the differentiable SLAM predicts a ground truth trajectory using them.

Classical SLAM systems~\cite{raul2015orbslam} are non-differentiable.
A common technique in these systems - the non-linear optimization is based on the Levenberg-Marquardt algorithm~\cite{madsen2004methods}. It switches the damping factor discretely at each iteration of the optimization process. This stops the gradient from backpropagating to the nodes when we unroll the optimization iterations to build the computational graph~\cite{gradslam}.
We use the generalized logistic function~\cite{richards1959flexible} for soft switching of the damping factor as well as the optimization update~\cite{gradslam} 
\begin{align}
\lambda = \lambda_{min} + \frac{\lambda_{max}-\lambda_{min}}{1 + D e^{-\sigma(r_{1}-r_{0})}},\\
x_{t+1} = x_t + \frac{\delta_t}{1 + e^{-(r_1 - r_0)}},
\end{align}

$\lambda_{\text{max}}$ and $\lambda_{\text{min}}$ are damping coefficient bounds in Levenberg-Marquardt solvers. $r_0$ and $r_1$ represent error norms at the current and lookahead iterates. $D$ and $\sigma$ are tunable parameters.
\section{Differentiable SLAM Integration}
\label{sec:arch}
We now discuss the methodology of integrating SLAM into DLMP training tasks.
In general, differentiable SLAM can be backpropagated as additional information in a deep learning model to help train the model for better performance.
To this end, we study three LiDAR-related tasks - Ground v/s non-Ground Segmentation, Ground Elevation Estimation, the Dynamic to Static LiDAR translation, and Generative modelling for LiDAR. 
\begin{figure*}[]
\begin{subfigure}{\textwidth 0pt}
  \centering
  \includegraphics[width=1\linewidth, height=0.25\linewidth]{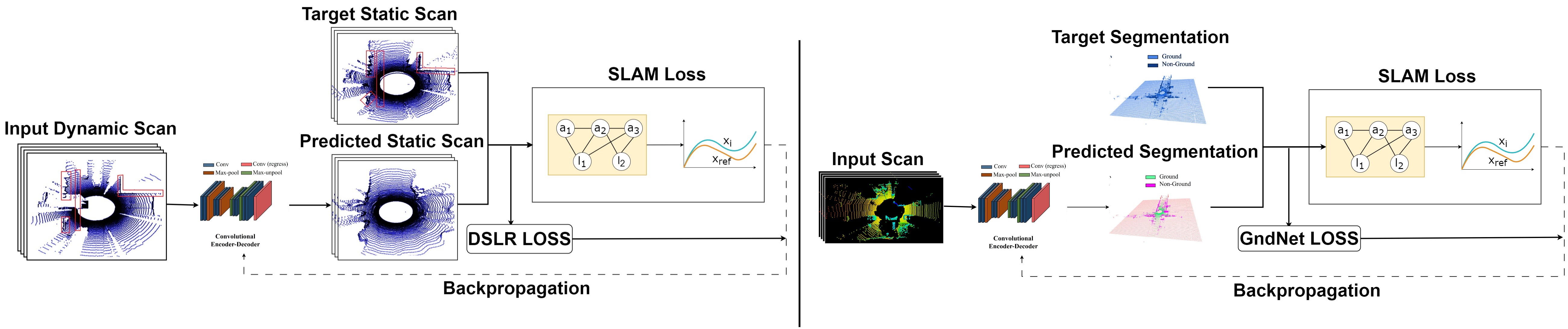} \caption{Left: Setup of DSLR with Differentiable SLAM Loss. Right: Setup of GndNet with Differentiable SLAM Loss.}
  \label{fig:gndnet-dslr}
\end{subfigure}
\end{figure*}
\subsection{Ground Elevation Estimation and Ground Segmentation}
\subsubsection{GndNet}
Ground Elevation Estimation for LiDAR scans is crucial for tasks like navigable space detection, registration, to name a few. GndNet~\cite{paigwar2020gndnet} estimates the ground elevation information as well as segments the LiDAR points into ground and non-ground (object/obstacle) points. We adapt their models to train with differentiable SLAM error along with the existing loss function. The goal is to achieve better estimates of ground plane elevation and classification into ground v/s non-ground points, using differentiable SLAM error.
GndNet discretizes the raw point cloud into a evenly spaced $\mathbf{x-y}$ grid, without binning the z-dimension (here the x, y, z direction refer to orientation of the LiDAR points in 3D coordinates), thereby creating a set of pillars~\cite{lang2019pointpillars}.
Next, PointNet~\cite{qi2017pointnet} is used to generate features for every non-empty pillar. 
Then, these pillar features are placed on the $\mathbf{x-y}$ grid leading to a psuedo-image. Finally, a convolutional encoder-decoder network learns features from this image and regresses the ground elevation per cell in the grid. \textit{This regression output is compared against the grountruth elevation to compute }
$L_{model}$.
Further based on the elevation, points above a threshold are classified as obstacle/object/above ground, while points below threshold are classified as ground points, thereby segmenting into ground and obstacle class.  
\subsubsection{Differentiable SLAM based GndNet} We insert our differentiable SLAM module after the regression of ground elevation per cell. Using the elevation output, we extract the corresponding points in the LiDAR scan that are classified as above ground (via a threshold parameter used by GndNet). We consider this as a predicted LiDAR scan ($\tilde{L}$ ) generated based on thresholding of the elevation output of the model.
Given that the dataset also has groundtruth elevation information, we generate a groundtruth LiDAR scan ($L$), by thresholding the LiDAR points with groundtruth elevation information (using the same threshold parameter).
Given we have a batch of contiguous groundtruth and predicted LiDAR using the above strategy, we can use the differentiable SLAM error module to estimate the trajectory for both the batches. Thereafter, we evaluate the rotational and translation trajectory error between both the trajectories ($L_{slam}$). This is our SLAM error that can be successfully backpropagated through the network owing to the differentiable SLAM module. For a visual description, refer Figure \ref{fig:gndnet-dslr}.
\subsection{Dynamic to Static LiDAR Translation}
\subsubsection{DSLR}
We choose a generative modelling application to show the effect of the differentiable SLAM on a generative modelling task.
Dynamic to Static Translation of LiDAR point cloud~\cite{kumar2021dynamic} translates a LiDAR scan with occlusions due to dynamic objects, to a fully static scan with all dynamic occlusions replaced by static background. \textit{DSLR} uses a 3-module based model - an Autoencoder, Pair Discriminator and an Advesarial Module to achieve the translation.
Given a set of dynamic scans $X=(X_1,X_2...X_n)$  and corresponding static scans $Y=(Y_1,Y_2...Y_n)$, the autoencoder simple learns to reconstructs a LiDAR scan. The pair discriminator module classsfies a LiDAR scan pair into 2 classes based on the below equation.
\begin{align}
&DI\left((l_{1}), \left(l_{2}\right)\right)=\left
\{\begin{array}
{ll}
1 &  \hspace{4pt} l_{1} \in X, l_{2} \in X  \\
0 &  \hspace{4pt} l_{1} \in X, l_{2} \in Y \\
\end{array}\right\}
\label{discequation}
\end{align}
The adversarial modules trick the discriminator to predict 1 for a pair that should be labelled as 0, thus generating the adversarial loss ($L_{model}$),  which helps to achieve static translation for a dynamic LiDAR scan.\\
\subsubsection{Differentiable SLAM based DSLR}
We modify the adversarial module of DSLR in order to plug differentiable SLAM.
Given a dynamic scan($d_i$) as input, the output of the adversarial module is a reconstructed static scan ($\bar{s_i}$) with the dynamic occlusions replaced by the actual static background. We also have the groundtruth static scans ($s_i$) to compare the generated scans against.
 Given that we have a batch of contiguous reconstructed static scans (($\bar{s_i}$)), as well as the groundtruth static scans ($s_i$), we can use the differentiable SLAM error module to calculate the trajectories for both the sets and calculate the error between the two ($L_{slam}$), that can be backpropagated using the deep learning model. For a visual description, refer Figure \ref{fig:gndnet-dslr}.

\subsection{Conditional LiDAR Generation ($AE_{lidar}$)}
We demonstrate the benefit of our differentiable SLAM module on a standard LiDAR autoencoder which serves as backbone for multiple downstream tasks. We use Caccia et. al. \cite{caccia2018deep} to reconstruct LiDAR scan using range-image based LiDAR representation. The encoder and decoder architectures are adapted from Radford et. al. \cite{radford2015unsupervised}. We train $AE_{lidar}$ with and without differentiable SLAM. For the differentiable SLAM variant, we calculate the SLAM loss between the reconstructed output LiDAR and the input LIDAR scan. The SLAM loss along with the reconstruction loss is back-propagated to ensure that the model learns from the SLAM error as well. The pipeline for this task is similar to DSLR (Figure \ref{fig:gndnet-dslr}(left)).

\section{Experiments}
\subsection{Experimental Setup}
\label{sec:expsetup}
For all the deep learning models used in our paper: DSLR\cite{kumar2021dynamic}, GndNet \cite{paigwar2020gndnet} and $AE_{lidar}$\cite{caccia2018deep} we follow the experimental setting of the respective models as used in their work, except a minor change for GndNet. GndNet uses every fourth contiguous LiDAR scan for training. However, we require finer contiguity as we compute SLAM error between contiguous scans. Therefore, we use every second contiguous scan for training GndNet and run the experiments with and without the SLAM module using this setting to report the results.

SLAM error module is time consuming. Thus, we do not calculate SLAM error for every epoch - we calculate SLAM error after every $k^{th}$ epoch, where k is a hyperparameter. More details in the  Appendix (Section \ref{training_appendix}).

\subsection{\textbf{Datasets}}

\textbf{CARLA-64}: CARLA-64 \cite{kumar2021dynamic} is an extensive simulated LiDAR dataset. It mimics the exact settings of a VLP-64 LiDAR  sensor. The dataset consists of 8 sequences for training and 6 for testing. It has 4 sequences for testing on SLAM.\\\textbf{ARD-16}: ARD-16 \cite{kumar2021dynamic} is a real-world sparse industrial dataset collected using a VLP-Puck LiDAR sensor. It is 4$\times$ sparse compared to CARLA-64 and KITTI. \\
\textbf{SemanticKITTI}: SemanticKITTI \cite{behley2019semantickitti}\cite{geiger2012we} is a well known LiDAR dataset with semantic segmentation labels. It has 11 sequences(00-10) for which semantic labels are available. \\For more details on the datasets, please refer to the Section \ref{dataset_appendix} in the Appendix.
\subsection{Results}
\subsubsection{}\textbf{Ground Elevation Estimation and Segmentation}
We, for the first time show the possibility of integration of a differentiable SLAM module into a regression task (ground elevation estimation) and segmentation (into ground v/s non-ground points) based downstream task.
We compare the results of the GndNet with and without differentiable SLAM error as explained in Section \ref{sec:arch}. The model performs 2 task - regressing the ground elevation of the scan points, and segmenting the points into ground and above ground points. The performance of the model is shown in the Table  \ref{tab:gndnet}. Our variant comfortably surpasses the MSE (Mean-Squared Error) estimate while  regressing the elevation of the points with an \textit{improvement of 0.04} on MSE.
Also our model fares better than GndNet on recall - increase of 3\% and  makes less false positive mistakes. This improvement is gained by using the differentiable SLAM module for only 27 epochs out of the total 150 epochs. (Refer \ref{sec:expsetup}) .
\begin{table}[htbp]
\centering
\begin{tabular}{|c|c|c|c|c|c|}
\hline
  \textbf{Method} & \textbf{Frames} & \textbf{MSE} & \textbf{mIOU} & \textbf{Prec} & \textbf{Recall}   \\ \hline
  
  GndNet & 6554 & 0.76 & \textbf{0.81} & \textbf{0.85} & 0.94\\
  \hline
  GndNet+Diff SLAM & 6554 & \textbf{0.72} & \textbf{0.81} & 0.83 & \textbf{0.97} \\
  \hline
\end{tabular}
\caption{Comparison of the Differentiable SLAM module on ground elevation estimation task and segmentation. For MSE, lower the better, for the rest, opposite.}
\label{tab:gndnet}
\end{table}
\begin{table}[]
    \centering
    \begin{tabular}{|c|c|c|c|}
    \hline
      \textbf{Dataset} & \textbf{Run} & \textbf{DSLR with Diff. SLAM} & \textbf{DSLR without Diff. SLAM} \\ 
      \hline
  
{}{}{CARLA-64} &
        
      Avg [9..14] & \textbf{6.96} & 7.85 \\
     \hline
     ARD-16 & 3 &   \textbf{0.31} & 0.34 \\ 
     \hline
     KITTI &  8 &   \textbf{5.00} & 5.23 \\ \hline
   
    \end{tabular}
    \caption{Relative Comparison of Static Translation(using Chamfer Distance) for DSLR with and without Differentiable SLAM. Lower the better.}
    \label{tab:chamfer}
\end{table}
\begin{figure}[h]
  \centering
  \includegraphics[width=0.8\linewidth, height=0.3\linewidth]{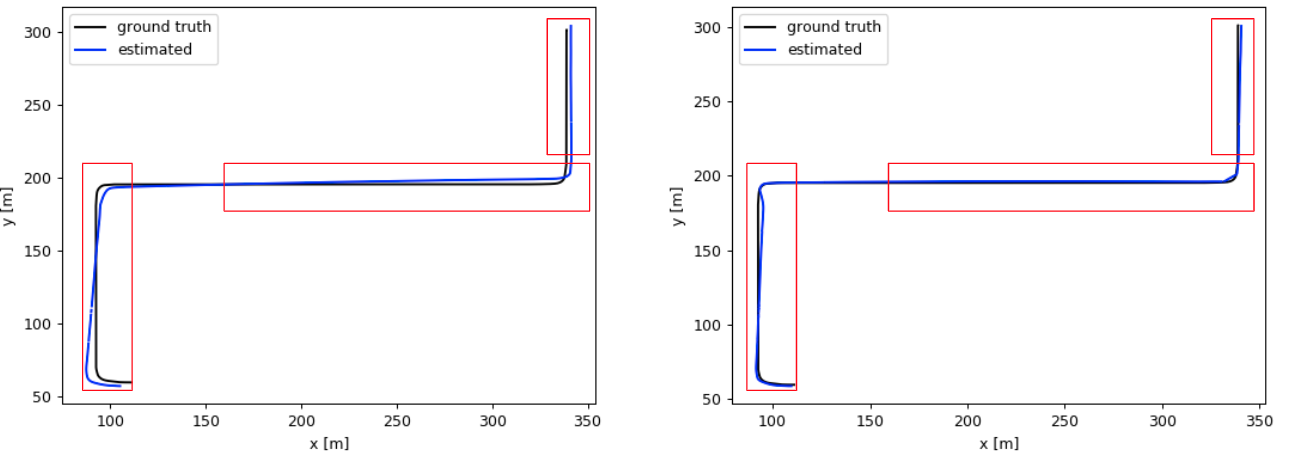}

  \caption{SLAM using \textbf{(1)} static output scans of DSLR(left)  and \textbf{(2)} with static output from DSLR+Differentiable SLAM Error model (right). Red boxes indicate regions where the trajectory generated from our framework is better than the baseline. Differentiable SLAM  helps to reduce the drift accumulated while navigation.}
  \label{fig:dslr-trajectory}
\end{figure}
\subsubsection{Dynamic to Static LiDAR Translation for SLAM}
In this section we for the first time show the application of differentiable SLAM for a generative modelling task - Dynamic to Static Translation for LiDAR scan for effective SLAM~\cite{kumar2021dynamic}.

In this task, we evaluate the relative benefit of using differentiable SLAM over plain DSLR.

We also compare the reconstruction quality of the static translations with the ground truth static scan using Chamfer's Distance\cite{emd-cd}. As we see in Table \ref{tab:chamfer},  with differentiable SLAM the Chamfer Distance is always better. Here we would like to point that we add SLAM error as a loss term only for 7 interleaved epochs (Section \ref{sec:expsetup}) which gives a meaningful reduction in the error. For results on all the six CARLA test sequences, please refer to Table \ref{tab:chamfer_appendix} in the Appendix. \\
We further investigate the effect of integrating the differentiable SLAM module in DSLR on downstream SLAM. As we see in Table \ref{tab:dslr-comparsion} and Figure \ref{fig:dslr-trajectory}, using static reconstructions obtained from differentiable SLAM integrated DSLR gives reduced navigation error on all the four LIDAR SLAM sequences for CARLA-64.

\begin{table}[]
\centering
\begin{tabular}{|ccccccc|}
\hline
\multicolumn{1}{|c|}{{}{}{Run}} & \multicolumn{3}{c|}{With Diff SLAM}                                                 & \multicolumn{3}{c|}{Without Diff SLAM}                                  \\ \cline{2-7} 
\multicolumn{1}{|c|}{}                     & \multicolumn{1}{c|}{ATE}  & \multicolumn{2}{c|}{RPE}                                & \multicolumn{1}{c|}{ATE}  & \multicolumn{2}{c|}{RPE}                    \\ \cline{2-7} 
\multicolumn{1}{|c|}{}                     & \multicolumn{1}{c|}{}     & \multicolumn{1}{c|}{Trans}   & \multicolumn{1}{c|}{Rot} & \multicolumn{1}{c|}{}     & \multicolumn{1}{c|}{Trans}   & Rot          \\ \hline
\multicolumn{7}{|c|}{CARLA-64 Dataset}                                                                                                                                                                     \\ \hline
\multicolumn{1}{|c|}{0}                    & \multicolumn{1}{c|}{\textbf{2.37}} & \multicolumn{1}{c|}{\textbf{0.440}} & \multicolumn{1}{c|}{\textbf{0.09}}  & \multicolumn{1}{c|}{4.73} & \multicolumn{1}{c|}{\textbf{0.440}}  & 0.11           \\ \hline

\multicolumn{1}{|c|}{1}                    & \multicolumn{1}{c|}{\textbf{1.3}}  & \multicolumn{1}{c|}{\textbf{0.400}}   & \multicolumn{1}{c|}{\textbf{0.070}}  & \multicolumn{1}{c|}{2.9}  & \multicolumn{1}{c|}{\textbf{0.400}}   & \textbf{0.070}           \\ \hline

\multicolumn{1}{|c|}{2}                    & \multicolumn{1}{c|}{\textbf{0.76}} & \multicolumn{1}{c|}{\textbf{0.567}} & \multicolumn{1}{c|}{\textbf{0.07}}  & \multicolumn{1}{c|}{1.36} & \multicolumn{1}{l|}{0.571} & 0.15           \\ \hline
\multicolumn{1}{|c|}{3}                    & \multicolumn{1}{c|}{\textbf{4.09}} & \multicolumn{1}{c|}{0.399} & \multicolumn{1}{c|}{\textbf{0.081}} & \multicolumn{1}{c|}{4.4}  & \multicolumn{1}{c|}{\textbf{0.395}} & 0.104          \\ \hline
\multicolumn{7}{|c|}{ARD-16 Dataset}                                                                                                                                                                       \\ \hline
\multicolumn{1}{|l|}{3}                    & \multicolumn{1}{c|}{\textbf{1.94}} & \multicolumn{1}{c|}{\textbf{4.81}}  & \multicolumn{1}{c|}{\textbf{0.186}} & \multicolumn{1}{c|}{2.05} & \multicolumn{1}{c|}{\textbf{4.81}}  & 0.188 \\ \hline
\end{tabular}
\caption{Relative Comparison of Static Translation 
for DSLR with and without Differentiable SLAM on CARLA-64 and ARD-16 dataset. We use Chamfer Distance metric.}
\label{tab:dslr-comparsion}
\end{table}

\subsubsection{LiDAR Reconstruction using $AE_{lidar}$}
\begin{table}[]
\begin{tabular}{|c|c|c|}
\hline
CARLA-64 & Chamfer Distance with SLAM & Chamfer Distance without SLAM \\ \hline
Avg[9..14] & \textbf{2.83} & 3.03 \\ \hline
\end{tabular}
\caption{Comparison of a general purpose generative autoencoder, $AE_{lidar}$ wiith and without differentiable SLAM using Chamfer's Distance metric.}
\label{tab:ae_lidar}
\end{table}
We demonstrate the effect of our differentiable SLAM on a general purpose simple generative model in Table \ref{tab:ae_lidar}. Our result demonstrates that such general purpose model that is used in several complex models as a \textit{backbone} can benefit from differentiable SLAM. For detailed results on all the CARLA test sequences, please refer to Table \ref{tab:ae_lidar_appendix} in the Appendix.
\section{Discussion and Limitations}
Certain limitations of Differentiable SLAM are discussed in the Appendix (Section \ref{limitation_appendix}).

LiDAR based applications has seen significant progress with the development of new techniques and technologies that have revolutionized the field. One such technique is SLAM, which is a popular approach to map an unknown environment and localize a robot within it. In this paper, we propose a novel method that uses differentiable SLAM to improve the performance of deep learning tasks such as binary segmentation, generative modeling, and regression.
Our core idea lies in the fact that SLAM prefers certain properties over others, such as static structures/non-ground points over dynamic/ground ones. We assume that the reference trajectory provided to SLAM is close to the ground truth, which enables us to minimize SLAM loss in a way that is equivalent to minimizing with ground truth poses. By doing so, we encourage DSLR to give more static-like predictions, and segmentation models to make clear distinctions between ground and non ground objects. Additionally, we use SLAM to improve elevation regression so that ground points can be deleted, which improves SLAM performance.
Our approach is based on a two-step reasoning process - we first assume that the reference trajectory provided to SLAM is close to the ground truth, and then we exploit the properties that SLAM prefers to improve deep learning tasks. We argue that SLAM preferences can be used to improve the performance of deep learning tasks. We present empirical results that demonstrate the effectiveness of our approach.
 Overall, we believe that our approach has the potential to significantly advance the field of robotics and open up new avenues for research in this exciting area.

\bibliography{egbib.bib}

\begin{thebibliography}{45}
\providecommand{\natexlab}[1]{#1}
\providecommand{\url}[1]{\texttt{#1}}
\expandafter\ifx\csname urlstyle\endcsname\relax
  \providecommand{\doi}[1]{doi: #1}\else
  \providecommand{\doi}{doi: \begingroup \urlstyle{rm}\Url}\fi

\bibitem[Behley et~al.(2019)Behley, Garbade, Milioto, Quenzel, Behnke, Stachniss, and Gall]{behley2019semantickitti}
Jens Behley, Martin Garbade, Andres Milioto, Jan Quenzel, Sven Behnke, Cyrill Stachniss, and Jurgen Gall.
\newblock Semantickitti: A dataset for semantic scene understanding of lidar sequences.
\newblock In \emph{Proceedings of the IEEE/CVF international conference on computer vision}, pages 9297--9307, 2019.

\bibitem[Bloembergen and Eijgenstein(2021)]{bloembergen2021automatic}
Daan Bloembergen and Chris Eijgenstein.
\newblock Automatic labelling of urban point clouds using data fusion.
\newblock \emph{arXiv preprint arXiv:2108.13757}, 2021.

\bibitem[Bloesch et~al.(2018)Bloesch, Czarnowski, Clark, Leutenegger, and Davison]{bloesch2018codeslam}
Michael Bloesch, Jan Czarnowski, Ronald Clark, Stefan Leutenegger, and Andrew~J. Davison.
\newblock Codeslam - learning a compact, optimisable representation for dense visual slam.
\newblock In \emph{2018 IEEE/CVF Conference on Computer Vision and Pattern Recognition}, pages 2560--2568, 2018.
\newblock \doi{10.1109/CVPR.2018.00271}.

\bibitem[{Caccia} et~al.(2019){Caccia}, v.~{Hoof}, {Courville}, and {Pineau}]{caccia2018deep}
L.~{Caccia}, H.~v.~{Hoof}, A.~{Courville}, and J.~{Pineau}.
\newblock Deep generative modeling of lidar data.
\newblock In \emph{2019 IEEE/RSJ International Conference on Intelligent Robots and Systems (IROS)}, pages 5034--5040, 2019.

\bibitem[Chen et~al.(2014)Chen, Lai, Wu, Martin, and Hu]{chen2014automatic}
Kang Chen, Yu-Kun Lai, Yu-Xin Wu, Ralph Martin, and Shi-Min Hu.
\newblock Automatic semantic modeling of indoor scenes from low-quality rgb-d data using contextual information.
\newblock \emph{ACM Transactions on Graphics}, 33\penalty0 (6), 2014.

\bibitem[Chen et~al.(2021)Chen, Li, Mersch, Wiesmann, Gall, Behley, and Stachniss]{chen2021moving}
Xieyuanli Chen, Shijie Li, Benedikt Mersch, Louis Wiesmann, J{\"u}rgen Gall, Jens Behley, and Cyrill Stachniss.
\newblock Moving object segmentation in 3d lidar data: A learning-based approach exploiting sequential data.
\newblock \emph{IEEE Robotics and Automation Letters}, 6\penalty0 (4):\penalty0 6529--6536, 2021.

\bibitem[Eskandar et~al.(2022)Eskandar, Palaniswamy, Guirguis, Somashekar, and Yang]{eskandar2022glpu}
George Eskandar, Janaranjani Palaniswamy, Karim Guirguis, Barath Somashekar, and Bin Yang.
\newblock Glpu: A geometric approach for lidar pointcloud upsampling.
\newblock \emph{arXiv preprint arXiv:2202.03901}, 2022.

\bibitem[FNU et~al.(2022)FNU, Vattikonda, Dong, and Sahoo]{fnu2022gradlidarslam}
Aryan FNU, Dheeraj Vattikonda, Erqun Dong, and Sabyasachi Sahoo.
\newblock Grad-lidar-{SLAM}: Fully differentiable global {SLAM} for lidar with pose-graph optimization.
\newblock In \emph{IROS 2022 Workshop Probabilistic Robotics in the Age of Deep Learning}, 2022.

\bibitem[Geiger et~al.(2012)Geiger, Lenz, and Urtasun]{geiger2012we}
Andreas Geiger, Philip Lenz, and Raquel Urtasun.
\newblock Are we ready for autonomous driving? the kitti vision benchmark suite.
\newblock In \emph{2012 IEEE conference on computer vision and pattern recognition}, pages 3354--3361. IEEE, 2012.

\bibitem[Guillard et~al.(2022)Guillard, Vemprala, Gupta, Miksik, Vineet, Fua, and Kapoor]{guillard2022learning}
Beno{\^\i}t Guillard, Sai Vemprala, Jayesh~K Gupta, Ondrej Miksik, Vibhav Vineet, Pascal Fua, and Ashish Kapoor.
\newblock Learning to simulate realistic lidars.
\newblock In \emph{2022 IEEE/RSJ International Conference on Intelligent Robots and Systems (IROS)}, pages 8173--8180. IEEE, 2022.

\bibitem[Hao-Su(2017)]{emd-cd}
Hao-Su.
\newblock 3d deep learning on point cloud representation (analysis), 2017.
\newblock URL \url{http://graphics.stanford.edu/courses/cs468-17-spring/LectureSlides/L14%20-%203d%20deep%20learning%20on%20point%20cloud%20representation%20(analysis).pdf}.

\bibitem[He et~al.(2020)He, Zeng, Huang, Hua, and Zhang]{he2020structure}
Chenhang He, Hui Zeng, Jianqiang Huang, Xian-Sheng Hua, and Lei Zhang.
\newblock Structure aware single-stage 3d object detection from point cloud.
\newblock In \emph{Proceedings of the IEEE/CVF conference on computer vision and pattern recognition}, pages 11873--11882, 2020.

\bibitem[Hu et~al.(2020)Hu, Yang, Xie, Rosa, Guo, Wang, Trigoni, and Markham]{hu2020randla}
Qingyong Hu, Bo~Yang, Linhai Xie, Stefano Rosa, Yulan Guo, Zhihua Wang, Niki Trigoni, and Andrew Markham.
\newblock Randla-net: Efficient semantic segmentation of large-scale point clouds.
\newblock In \emph{Proceedings of the IEEE/CVF conference on computer vision and pattern recognition}, pages 11108--11117, 2020.

\bibitem[Jatavallabhula et~al.(2020)Jatavallabhula, Iyer, and Paull]{gradslam}
Krishna~Murthy Jatavallabhula, Ganesh Iyer, and Liam Paull.
\newblock Gradslam: Dense slam meets automatic differentiation.
\newblock In \emph{2020 IEEE International Conference on Robotics and Automation (ICRA)}, pages 2130--2137, 2020.
\newblock \doi{10.1109/ICRA40945.2020.9197519}.

\bibitem[Kim et~al.(2020)Kim, Yoo, and Jung]{kim2020color}
Hyun-Koo Kim, Kook-Yeol Yoo, and Ho-Youl Jung.
\newblock Color image generation from range and reflection data of lidar.
\newblock \emph{Sensors}, 20\penalty0 (18):\penalty0 5414, 2020.

\bibitem[Kumar et~al.(2021)Kumar, Sahoo, Shah, Kondameedi, Jain, Verma, Bhattacharyya, and Vishwanath]{kumar2021dynamic}
Prashant Kumar, Sabyasachi Sahoo, Vanshil Shah, Vineetha Kondameedi, Abhinav Jain, Akshaj Verma, Chiranjib Bhattacharyya, and Vinay Vishwanath.
\newblock Dynamic to static lidar scan reconstruction using adversarially trained auto encoder.
\newblock In \emph{Proceedings of the AAAI Conference on Artificial Intelligence}, volume~35, pages 1836--1844, 2021.

\bibitem[Landrieu and Simonovsky(2018)]{landrieu2018large}
Loic Landrieu and Martin Simonovsky.
\newblock Large-scale point cloud semantic segmentation with superpoint graphs.
\newblock In \emph{Proceedings of the IEEE conference on computer vision and pattern recognition}, pages 4558--4567, 2018.

\bibitem[Lang et~al.(2019)Lang, Vora, Caesar, Zhou, Yang, and Beijbom]{lang2019pointpillars}
Alex~H Lang, Sourabh Vora, Holger Caesar, Lubing Zhou, Jiong Yang, and Oscar Beijbom.
\newblock Pointpillars: Fast encoders for object detection from point clouds.
\newblock In \emph{Proceedings of the IEEE/CVF conference on computer vision and pattern recognition}, pages 12697--12705, 2019.

\bibitem[Lee et~al.(2022)Lee, Lim, and Myung]{lee2022patchwork++}
Seungjae Lee, Hyungtae Lim, and Hyun Myung.
\newblock Patchwork++: Fast and robust ground segmentation solving partial under-segmentation using 3d point cloud.
\newblock In \emph{2022 IEEE/RSJ International Conference on Intelligent Robots and Systems (IROS)}, pages 13276--13283. IEEE, 2022.

\bibitem[Lim et~al.(2021)Lim, Oh, and Myung]{lim2021patchwork}
Hyungtae Lim, Minho Oh, and Hyun Myung.
\newblock Patchwork: Concentric zone-based region-wise ground segmentation with ground likelihood estimation using a 3d lidar sensor.
\newblock \emph{IEEE Robotics and Automation Letters}, 6\penalty0 (4):\penalty0 6458--6465, 2021.

\bibitem[Lu et~al.(2022)Lu, Zhang, Doherty, Severinsen, Yang, and Leonard]{lu2022slam}
Ziqi Lu, Yihao Zhang, Kevin Doherty, Odin Severinsen, Ethan Yang, and John Leonard.
\newblock Slam-supported self-training for 6d object pose estimation.
\newblock In \emph{2022 IEEE/RSJ International Conference on Intelligent Robots and Systems (IROS)}, pages 2833--2840. IEEE, 2022.

\bibitem[Madsen et~al.(2004)Madsen, Nielsen, and Tingleff]{madsen2004methods}
Kaj Madsen, Hans~Bruun Nielsen, and Ole Tingleff.
\newblock Methods for non-linear least squares problems.
\newblock 2004.

\bibitem[Merrill et~al.(2022)Merrill, Guo, Zuo, Huang, Leutenegger, Peng, Ren, and Huang]{merrill2022symmetry}
Nathaniel Merrill, Yuliang Guo, Xingxing Zuo, Xinyu Huang, Stefan Leutenegger, Xi~Peng, Liu Ren, and Guoquan Huang.
\newblock Symmetry and uncertainty-aware object slam for 6dof object pose estimation.
\newblock In \emph{Proceedings of the IEEE/CVF Conference on Computer Vision and Pattern Recognition}, pages 14901--14910, 2022.

\bibitem[Milioto et~al.(2019)Milioto, Vizzo, Behley, and Stachniss]{milioto2019rangenet++}
Andres Milioto, Ignacio Vizzo, Jens Behley, and Cyrill Stachniss.
\newblock Rangenet++: Fast and accurate lidar semantic segmentation.
\newblock In \emph{2019 IEEE/RSJ international conference on intelligent robots and systems (IROS)}, pages 4213--4220. IEEE, 2019.

\bibitem[Mur-Artal et~al.(2015)Mur-Artal, Montiel, and Tard\'os]{raul2015orbslam}
Ra\'ul Mur-Artal, J.~M.~M. Montiel, and Juan~D. Tard\'os.
\newblock {ORB-SLAM}: a versatile and accurate monocular {SLAM} system.
\newblock \emph{IEEE Transactions on Robotics}, 31\penalty0 (5):\penalty0 1147--1163, 2015.
\newblock \doi{10.1109/TRO.2015.2463671}.

\bibitem[Nakashima and Kurazume(2021)]{nakashima2021learning}
Kazuto Nakashima and Ryo Kurazume.
\newblock Learning to drop points for lidar scan synthesis.
\newblock In \emph{2021 IEEE/RSJ International Conference on Intelligent Robots and Systems (IROS)}, pages 222--229. IEEE, 2021.

\bibitem[Nakashima et~al.(2023)Nakashima, Iwashita, and Kurazume]{nakashima2023generative}
Kazuto Nakashima, Yumi Iwashita, and Ryo Kurazume.
\newblock Generative range imaging for learning scene priors of 3d lidar data.
\newblock In \emph{Proceedings of the IEEE/CVF Winter Conference on Applications of Computer Vision}, pages 1256--1266, 2023.

\bibitem[Paigwar et~al.(2020)Paigwar, Erkent, Sierra-Gonzalez, and Laugier]{paigwar2020gndnet}
Anshul Paigwar, Özgür Erkent, David Sierra-Gonzalez, and Christian Laugier.
\newblock Gndnet: Fast ground plane estimation and point cloud segmentation for autonomous vehicles.
\newblock In \emph{2020 IEEE/RSJ International Conference on Intelligent Robots and Systems (IROS)}, pages 2150--2156, 2020.
\newblock \doi{10.1109/IROS45743.2020.9340979}.

\bibitem[Qi et~al.(2017)Qi, Su, Mo, and Guibas]{qi2017pointnet}
Charles~R Qi, Hao Su, Kaichun Mo, and Leonidas~J Guibas.
\newblock Pointnet: Deep learning on point sets for 3d classification and segmentation.
\newblock In \emph{Proceedings of the IEEE conference on computer vision and pattern recognition}, pages 652--660, 2017.

\bibitem[Qi et~al.(2018)Qi, Liu, Wu, Su, and Guibas]{qi2018frustum}
Charles~R Qi, Wei Liu, Chenxia Wu, Hao Su, and Leonidas~J Guibas.
\newblock Frustum pointnets for 3d object detection from rgb-d data.
\newblock In \emph{Proceedings of the IEEE conference on computer vision and pattern recognition}, pages 918--927, 2018.

\bibitem[Radford et~al.(2015)Radford, Metz, and Chintala]{radford2015unsupervised}
Alec Radford, Luke Metz, and Soumith Chintala.
\newblock Unsupervised representation learning with deep convolutional generative adversarial networks.
\newblock \emph{arXiv preprint arXiv:1511.06434}, 2015.

\bibitem[Richards(1959)]{richards1959flexible}
Francis~J Richards.
\newblock A flexible growth function for empirical use.
\newblock \emph{Journal of experimental Botany}, 10\penalty0 (2):\penalty0 290--301, 1959.

\bibitem[Shi et~al.(2019)Shi, Wang, and Li]{shi2019pointrcnn}
Shaoshuai Shi, Xiaogang Wang, and Hongsheng Li.
\newblock Pointrcnn: 3d object proposal generation and detection from point cloud.
\newblock In \emph{Proceedings of the IEEE/CVF conference on computer vision and pattern recognition}, pages 770--779, 2019.

\bibitem[Shi et~al.(2020)Shi, Wang, Shi, Wang, and Li]{shi2020points}
Shaoshuai Shi, Zhe Wang, Jianping Shi, Xiaogang Wang, and Hongsheng Li.
\newblock From points to parts: 3d object detection from point cloud with part-aware and part-aggregation network.
\newblock \emph{IEEE transactions on pattern analysis and machine intelligence}, 43\penalty0 (8):\penalty0 2647--2664, 2020.

\bibitem[Sodhi et~al.(2022)Sodhi, Dexheimer, Mukadam, Anderson, and Kaess]{sodhi2022leo}
Paloma Sodhi, Eric Dexheimer, Mustafa Mukadam, Stuart Anderson, and Michael Kaess.
\newblock Leo: Learning energy-based models in factor graph optimization.
\newblock In \emph{Conference on Robot Learning}, pages 234--244. PMLR, 2022.

\bibitem[Tateno et~al.(2017)Tateno, Tombari, Laina, and Navab]{Tateno2017CNNSLAMRD}
Keisuke Tateno, Federico Tombari, Iro Laina, and Nassir Navab.
\newblock Cnn-slam: Real-time dense monocular slam with learned depth prediction.
\newblock \emph{2017 IEEE Conference on Computer Vision and Pattern Recognition (CVPR)}, pages 6565--6574, 2017.

\bibitem[Triess et~al.(2022)Triess, Rist, Peter, and Z{\"o}llner]{triess2022realism}
Larissa~T Triess, Christoph~B Rist, David Peter, and J~Marius Z{\"o}llner.
\newblock A realism metric for generated lidar point clouds.
\newblock \emph{International Journal of Computer Vision}, 130\penalty0 (12):\penalty0 2962--2979, 2022.

\bibitem[Yan et~al.(2018)Yan, Mao, and Li]{yan2018second}
Yan Yan, Yuxing Mao, and Bo~Li.
\newblock Second: Sparsely embedded convolutional detection.
\newblock \emph{Sensors}, 18\penalty0 (10):\penalty0 3337, 2018.

\bibitem[Yang et~al.(2020)Yang, Sun, Liu, and Jia]{yang20203dssd}
Zetong Yang, Yanan Sun, Shu Liu, and Jiaya Jia.
\newblock 3dssd: Point-based 3d single stage object detector.
\newblock In \emph{Proceedings of the IEEE/CVF conference on computer vision and pattern recognition}, pages 11040--11048, 2020.

\bibitem[Yi et~al.(2021)Yi, Lee, Kloss, Mart{\'\i}n-Mart{\'\i}n, and Bohg]{yi2021differentiable}
Brent Yi, Michelle~A Lee, Alina Kloss, Roberto Mart{\'\i}n-Mart{\'\i}n, and Jeannette Bohg.
\newblock Differentiable factor graph optimization for learning smoothers.
\newblock In \emph{2021 IEEE/RSJ International Conference on Intelligent Robots and Systems (IROS)}, pages 1339--1345. IEEE, 2021.

\bibitem[Yin et~al.(2021)Yin, Zhou, and Krahenbuhl]{yin2021center}
Tianwei Yin, Xingyi Zhou, and Philipp Krahenbuhl.
\newblock Center-based 3d object detection and tracking.
\newblock In \emph{Proceedings of the IEEE/CVF conference on computer vision and pattern recognition}, pages 11784--11793, 2021.

\bibitem[Zhang et~al.(2020)Zhang, Zhou, David, Yue, Xi, Gong, and Foroosh]{zhang2020polarnet}
Yang Zhang, Zixiang Zhou, Philip David, Xiangyu Yue, Zerong Xi, Boqing Gong, and Hassan Foroosh.
\newblock Polarnet: An improved grid representation for online lidar point clouds semantic segmentation.
\newblock In \emph{Proceedings of the IEEE/CVF Conference on Computer Vision and Pattern Recognition}, pages 9601--9610, 2020.

\bibitem[Zhi et~al.(2019)Zhi, Bloesch, Leutenegger, and Davison]{Zhi2019SceneCodeMD}
Shuaifeng Zhi, Michael Bloesch, Stefan Leutenegger, and Andrew~J. Davison.
\newblock Scenecode: Monocular dense semantic reconstruction using learned encoded scene representations.
\newblock \emph{2019 IEEE/CVF Conference on Computer Vision and Pattern Recognition (CVPR)}, pages 11768--11777, 2019.

\bibitem[Zhu et~al.(2020)Zhu, Ma, Wang, Xu, Shi, and Lin]{zhu2020ssn}
Xinge Zhu, Yuexin Ma, Tai Wang, Yan Xu, Jianping Shi, and Dahua Lin.
\newblock Ssn: Shape signature networks for multi-class object detection from point clouds.
\newblock In \emph{Computer Vision--ECCV 2020: 16th European Conference, Glasgow, UK, August 23--28, 2020, Proceedings, Part XXV 16}, pages 581--597. Springer, 2020.

\bibitem[Zyrianov et~al.(2022)Zyrianov, Zhu, and Wang]{zyrianov2022learning}
Vlas Zyrianov, Xiyue Zhu, and Shenlong Wang.
\newblock Learning to generate realistic lidar point clouds.
\newblock In \emph{Computer Vision--ECCV 2022: 17th European Conference, Tel Aviv, Israel, October 23--27, 2022, Proceedings, Part XXIII}, pages 17--35. Springer, 2022.

\end{thebibliography}
\clearpage

\section{Appendix}
\subsection{Datasets}
 \label{dataset_appendix}
\textbf{CARLA-64}: CARLA-64 \cite{kumar2021dynamic} is an extensive simulated dataset. It mimics the exact settings of a VLP-64 LiDAR sensor. The dataset consists of 15 bunch of LiDAR scans (2048 scans per bunch), 00-07 for testing, 08 for validation while the and 09-14 for testing. A point to note here is that these do not have groundtruth poses for SLAM. The dataset provides with four separate test run for testing navigation performance.\\
We use this dataset for training DSLR, since it consists of paired static-dynamic correspondence for LiDAR scans which is required to train DSLR and measure the quality of the translations.

\textbf{ARD-16}: ARD-16 \cite{kumar2021dynamic} is a real-world sparse industrial dataset collected using a VLP-Puck LiDAR sensor. It is 4$\times$ sparse to the other two datasets. ARD-16 has paired correspondence available. We used it for testing our differentiable SLAM module for Dynamic to Static Translation.

\textbf{SemanticKITTI dataset}: We train and evaluate  GndNet and DSLR on the SemanticKITTI dataset \cite{behley2019semantickitti}\cite{geiger2012we}. It has 11 sequences(00-10) for which semantic labels are available. We follow the training and testing protocol of GndNet w.r.t SemanticKITTI. We use sequences 00 ,02 ,03 ,04 ,06 ,08 ,10 for training and 01 ,05 ,07 ,09 for test. We have 8323 scans for training and 6554 for testing. For DSLR, we train on all the sequences except 08, which is used for testing.

\subsection{Training Details}
\label{training_appendix}
For training the deep learning models we use a warmup of some epochs, after which we use the differentiable SLAM module once every $k$ epochs. \textit{k} is set to 5 for GndNet and DSLR and 10 for $AE_{lidar}.$

For \textbf{DSLR}, we train the adversarial module with differentiable SLAM error. Given we train the adversarial module from scratch, we use a warmup of 15 epochs. After the warmup, we use the SLAM error after every 5 epochs, as in the case of GndNet. Thus out of the 50 epochs for which the model is trained, differentiable SLAM is used in 7 epochs. We use a 24 GB RTX 3090 GPU for training.

\textbf{GndNet} takes only 6 hours to train without SLAM, and about a day with SLAM Error. We train it from scratch using the SLAM error module. We use a warmup of 15 epochs because the ground elevation estimation gives highly erroneous prediction at the start of training, which leads to inaccurate LiDAR scan generated from the predictions at the start. After the warmup, we use the SLAM error module \textit{once every 5 epochs}. This is because the differnetiable SLAM module is computationally expensive and takes 1 hours to train per epoch.

\subsection{Results}

We give detailed results on all the sequences of CARLA on two tasks - Dynamic to Static Translation using DSLR and Generative Modelling using an Autoencoder in Table \ref{tab:chamfer_appendix}
 and \ref{tab:ae_lidar_appendix}. Using differentiable SLAM with these two tasks helps to generate better results for both the tasks on majority of the CARLA sequences.
\begin{table}[htbp]
    \centering
    \begin{tabular}{|c|c|c|c|}
    \hline
      Dataset & \textbf{Run} & \textbf{DSLR with Diff. SLAM} & \textbf{DSLR without Diff. SLAM} \\ 
      \Xhline{3\arrayrulewidth}
  
{}{}{CARLA-64} &
        9 & \textbf{4.15} & 4.24 \\ \cline{2-4}
     & 10 & \textbf{14.55} & 16.24 \\ \cline{2-4}
     & 11 & \textbf{6.22} & 7.63 \\\cline{2-4}
     & 12 & 4.63 & \textbf{4.45} \\\cline{2-4}
     & 13 & \textbf{6.62} &  8.20\\\cline{2-4}
     & 14 & \textbf{5.59} & 6.31 \\\cline{2-4} 
     \Xhline{3\arrayrulewidth}
     ARD-16 & 3 &   \textbf{0.31} & 0.34 \\ 
     \Xhline{3\arrayrulewidth}
     KITTI &  8 &   \textbf{5.00} & 5.23 \\ \hline
   
    \end{tabular}
    \caption{Relative Comparison of Static Translation(using Chamfer Distance(Refer Supp.) ) for DSLR with and without Differentiable SLAM.}
    \label{tab:chamfer_appendix}
\end{table}.

\begin{table}[htbp]
\begin{tabular}{|c|c|c|}
\hline
\textbf{CARLA-64} & \textbf{Chamfer's Distance with SLAM} & \textbf{Chamfer's Distance without SLAM} \\ \hline
8                 & \textbf{2.1}                       & 2.28                                   \\ \hline
9                 & \textbf{1.58}                      & 1.91                                   \\ \hline
10                & \textbf{3.69}                      & 4.57                                   \\ \hline
11                & \textbf{3.01}                      & 3.35                                   \\ \hline
12                & 1.78                               & \textbf{1.31}                          \\ \hline
13                & \textbf{3.11}                      & 3.9                                    \\ \hline
14                & 4.55                               & \textbf{3.92}                          \\ \hline
\end{tabular}
\caption{Comparison of a general purpose generative autoencoder, $AE_{lidar}$ with and without differentiable SLAM using Chamfer's Distance metric.}
\label{tab:ae_lidar_appendix}
\end{table}

\subsection{Limitations}
\label{limitation_appendix}
One of the major limitations we face is that integration of differentiable SLAM with multi-class per point classification based applications (eg. semantic segmentation) is not differentiable as discussed in Section \ref{related-2}. This is a bottleneck towards many important applications and further research needs to be conducted here.
Another limitation is that the SLAM loss calculation in the differentiable SLAM module is time-consuming. More research is required for optimization of the SLAM error calculation module in the differentiable SLAM architecture to ensure faster training. Additionally we use gradSLAM, a local SLAM algorithm, which has been shown to suffer from high ATE due to the accumulation of drift over time due to absence of loop closure constraints. Unlike global SLAM algorithms, local SLAM algorithms do not correct for accumulated drift using loop closures. Future work on integrating loop closure constraints can help to strengthen the differentiable SLAM module.

\end{document}